\UseRawInputEncoding
\documentclass[conference]{IEEEtran}
\IEEEoverridecommandlockouts
\usepackage{cite}
\usepackage{amsmath,amssymb,amsfonts}
\usepackage{algorithmic}
\usepackage{graphicx}
\usepackage{textcomp}
\usepackage{xcolor}
\usepackage{fontawesome5}
\usepackage{hyperref}
\usepackage{multirow}

\makeatletter
\newcommand{\linebreakand}{%
  \end{@IEEEauthorhalign}
  \hfill\mbox{}\par
  \mbox{}\hfill\begin{@IEEEauthorhalign}
}
\makeatother

\def\BibTeX{{\rm B\kern-.05em{\sc i\kern-.025em b}\kern-.08em
    T\kern-.1667em\lower.7ex\hbox{E}\kern-.125emX}}

\begin{document}

\title{Large-Scale Traffic Congestion Prediction based on Multimodal Fusion and Representation Mapping\\}

\author{\IEEEauthorblockN{Bodong Zhou$^*$}
\IEEEauthorblockA{luckkyzhou@gmail.com \\ \textit{Individual Researcher}}
\and
\IEEEauthorblockN{Jiahui Liu$^{* \dagger}$}
\IEEEauthorblockA{
jliu1@connect.hku.hk \\ \textit{The University of Hong Kong}}
\and
\IEEEauthorblockN{Songyi Cui}
\IEEEauthorblockA{
echo07@connect.hku.hk \\ \textit{The University of Hong Kong}}
\and
\IEEEauthorblockN{Yaping Zhao$^\dagger$}
\IEEEauthorblockA{
zhaoyp@connect.hku.hk \\ \textit{The University of Hong Kong}}
\linebreakand
\IEEEauthorblockN{$^*$ Joint first authors.}
\and
\IEEEauthorblockN{$^\dagger$ Corresponding authors.}
}

\maketitle

\begin{abstract}

With the progress of the urbanisation process, the urban transportation system is extremely critical to the development of cities and the quality of life of the citizens. Among them, it is one of the most important tasks to judge traffic congestion by analysing the congestion factors. Recently, various traditional and machine-learning-based models have been introduced for predicting traffic congestion. However, these models are either poorly aggregated for massive congestion factors or fail to make accurate predictions for every precise location in large-scale space. To alleviate these problems, a novel end-to-end framework based on convolutional neural networks is proposed in this paper. With learning representations, the framework proposes a novel multimodal fusion module and a novel representation mapping module to achieve traffic congestion predictions on arbitrary query locations on a large-scale map, combined with various global reference information. The proposed framework achieves significant results and efficient inference on real-world large-scale datasets. 

\end{abstract}


\section{Introduction}

With the rapid development of the automobile industry in recent decades, the production and ownership of motor vehicles have increased significantly. At the same time, the increasing traffic demand is imposed on the urban transportation system due to the accelerated urbanisation process. Especially during peak hours, the road capacity of some road sections cannot meet the huge traffic demand, resulting in congestion problems. Severe traffic congestion may lead to extra carbon emissions and reduce the transportation network efficiency, even bring huge economic losses~\cite{fou2017}. Therefore, it has always been regarded as one of the most important traffic problems by analysing the factors of congestion and establishing traffic congestion prediction. Traffic congestion prediction plays an important role in a wide range of applications where: 1) Advanced Traffic Management Systems (ATMSs) and Advanced Traveller Information Systems provide real-time guidance information to travellers; 2) individuals or authorities optimise resources allocation regarding the required traffic time to ensure the journey smooth for travellers; 3) governments and urban planners design road network expansion plans~\cite{zheng2006}.

In the early decades, due to the lack of data availability, traffic prediction was biased towards predicting the traffic parameters of individual roads and local networks. Researchers at that time preferred to apply classical statistical models to predict traffic volumes. However, with the development of traffic, roads tend to be networked, and traditional statistical models can not well address the complexity of traffic data~\cite{cheng2018deeptransport}. However, the rapid advancement of artificial intelligence and machine learning has empowered a new wave of evolutionary solutions to many previously intractable engineering problems. Machine learning models can automatically adjust predictive model parameters to compensate for the shortcomings of traditional models, and thus are widely used in the field of traffic flow prediction. At present, the mainstream prediction models based on machine learning are Recurrent Neural Network (RNN), Convolutional Neural Network(CNN), and Factorisation Machine supported Neural Network (FNN)~\cite{ted2020}.

\begin{figure}[t]
\centerline{\includegraphics[width= 0.49\textwidth]{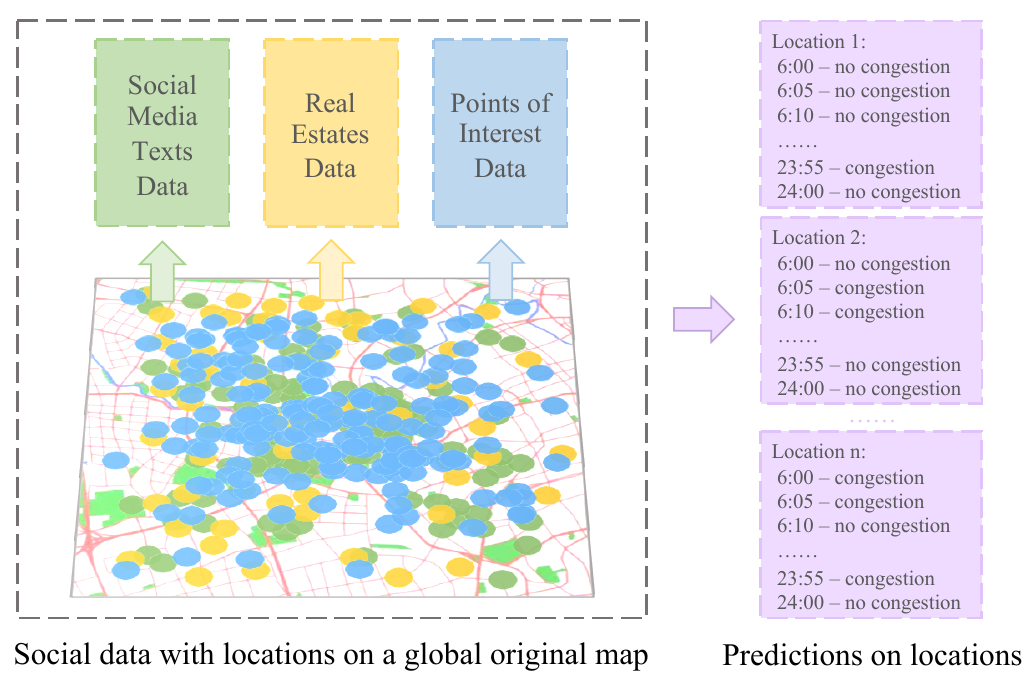}}
\caption{A variety of multimodal information is created at different locations on a large-scale map of the real world, and any localised traffic situation can be learned based on these diverse information.}
\label{intro}
\end{figure}

As one of the classical neural network models, FNN's fully connected structure enables it to process all functional combinations of the previous layer, but this connection structure also makes the training process very time-consuming~\cite{cheng2018deeptransport}. The RNN-based LSTM model has achieved very good results in the field of time series forecasting of traffic flow forecasting models. However, when the input data is a high-resolution two-dimensional data sequence, the processing speed of LSTM is slow~\cite{xie2019sequential}. Unlike RNN and FNN, CNN does not have any difficulty in handling high-resolution data, and its local connection structure also makes the training process shorter than the other two methods.

While the availability of big data permits researchers to explore this field further, different factors that affect traffic congestion, \textit{e.g.}, social media, economic state, and urban layout elements, are not fully considered. As a typical complex system of the urban road network, reasonably considering the interaction between different factors can effectively improve prediction accuracy. Also, incorporating diverse factors from the above-mentioned aspects for traffic congestion prediction is still unexplored. Therefore, aiming at traffic congestion prediction, designing a complex framework based on multimodal fusion and representation mapping becomes urgent and necessary with time as larger datasets can be accessed.

In order to better fuse the rich information on a large-scale map and infer traffic situations, a novel and efficient framework is proposed in this paper so that the global multimodal information referenced by traffic prediction is aggregated into a geo-preserving representation in a high-dimensional superspace, which can be utilised with mapped locations to predict traffic situations at different locations and time points (Fig.~\ref{intro}).

The main contributions of this paper are summarised as follows:

\begin{itemize}

\item A novel and lightweight end-to-end traffic congestion prediction framework based on representation fusion, which can use diverse multimodal global information from a large-scale map as the reference to predict traffic congestion at precise geographic locations and different time slots efficiently.

\item A global meta-representation learning method that generalises the multimodal global reference information on the corresponding spatial geographic structure of a large-scale map, learned by a Multimodal Fusion and Generalisation Module.

\item A geographical query mechanism based on a Query Location Mapping Module, which enables efficient traffic congestion inference through learning and mapping location representations with the module.

\item Intensive experiments and comprehensive analysis on real-world large-scale geolocated datasets, which demonstrate the effectiveness of the proposed method and stimulate further applications for road traffic planning.

\end{itemize}

\section{Related Work}

\subsection{Traffic Congestion Prediction}
The field of traffic forecasting has existed for nearly fifty years, and most of the initial research originated from civil engineering and traffic engineering research. However, with the rise of supercomputers and artificial intelligence technology, traffic flow prediction has gradually become interdisciplinary research. Mainstream traffic flow forecasting models are mainly divided into three categories. The first category belongs to classical statistical models, of which the Auto Regressive Integrated Moving Average (ARIMA) family of models is the most widely used~\cite{ted2020}. ARIMA was first applied in transportation prediction in 1979~\cite{ahmed1979analysis}. It was then applied to traffic flow prediction on highways by Levin and Tsao and reached the conclusion with the highest statistical significance of ARIMA(0,1,1)~\cite{levin1980forecasting}. With the development of urban traffic, traffic data tends to be nonlinear and big data, and traditional statistical models are no longer applicable~\cite{karlaftis2011statistical, li2018brief}.

\begin{figure}[t]
\centerline{\includegraphics[width= 0.49\textwidth]{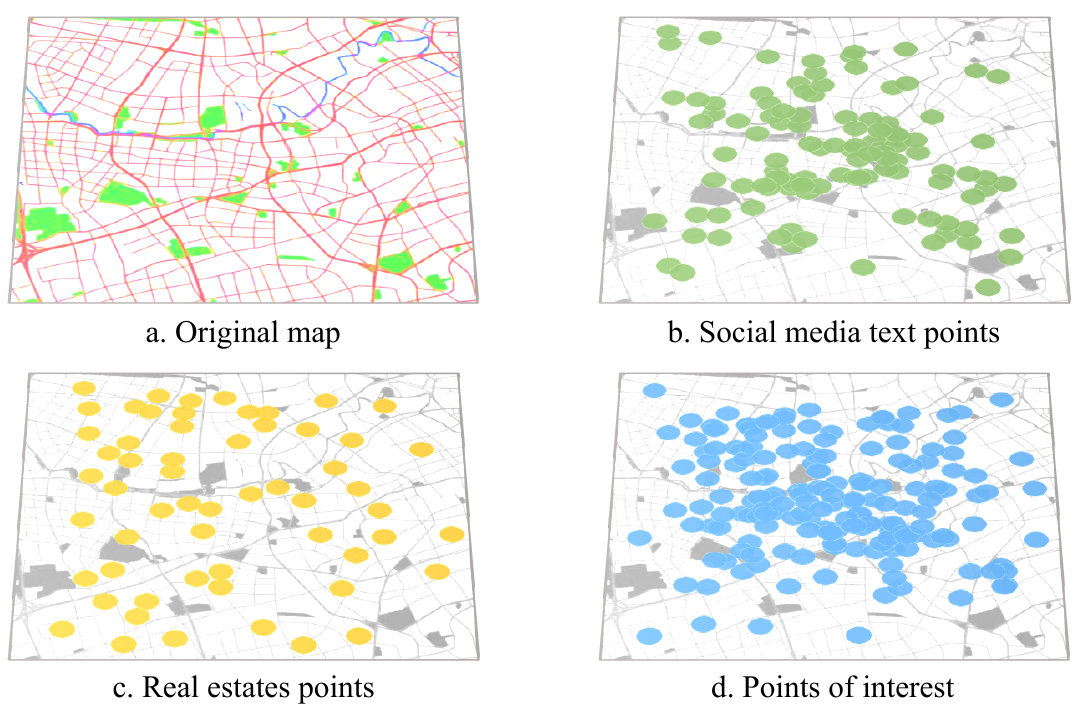}}
\caption{Social media texts, real estates, and points of interest are geographically scattered on a large-scale map with different distributions.}
\label{data}
\end{figure}

\begin{figure*}[t]
\centerline{\includegraphics[width= 1.0\textwidth]{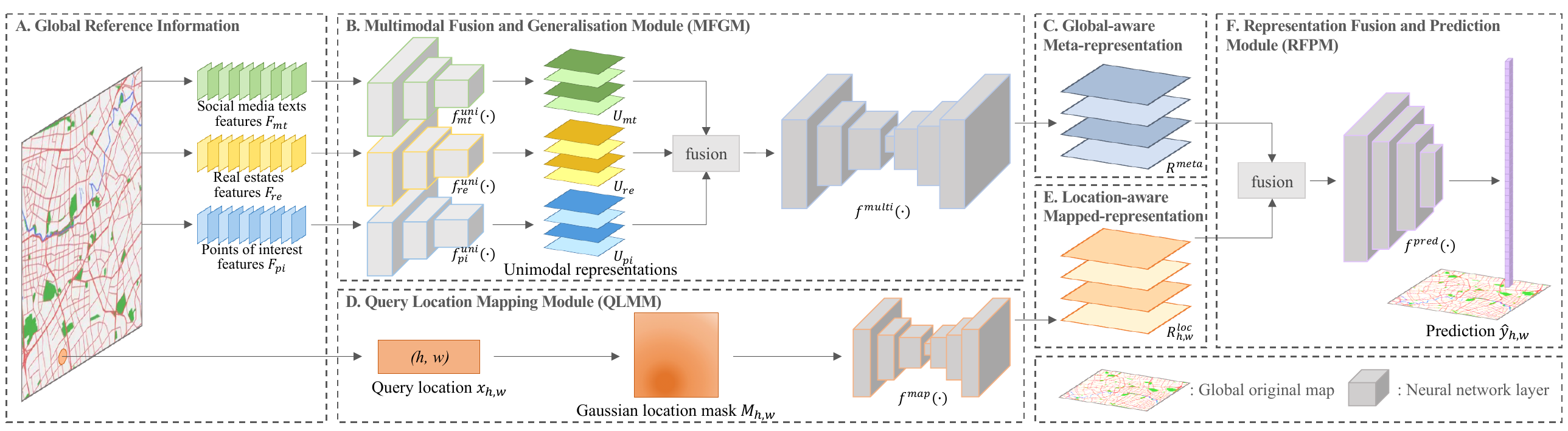}}
\caption{The proposed framework. Firstly, the featured Global Reference Information is fed into the Multimodal Fusion and Generalisation Module (MFGM) to obtain a Global-aware Meta-representation in a high-dimensional superspace. Also, the Query Location Mapping Module (QLMM) maps a query location as a Location-aware Mapped-representation in the same superspace. Finally, the Representation Fusion and Prediction Module (RFPM) makes prediction on traffic congestion based on the Global-aware Meta-representation and Location-aware Mapped-representation.}
\label{framework}
\end{figure*}

Due to the shortcomings of classical statistical models, the researchers start to pay attention to non-parametric models, k-nearest neighbour (KNN), support vector machine (SVM), Bayesian network (BN), and artificial neural network (ANN)~\cite{27, 28, 32, 34}. Machine learning is gradually applied to the field of traffic flow prediction. Chan et al. proposed a new neural network training method using a hybrid exponential smoothing method and the Levenberg-Marquardt (LM) algorithm, aiming to improve the generalisation ability of traffic flow prediction methods previously used to train neural networks. The improved neural network training method can denoise the traffic flow data~\cite{chan2011neural}. Cai et al. proposed an improved KNN model to improve prediction accuracy based on spatiotemporal correlations, enabling multi-step prediction. Unlike the previous KNN model that uses time series to describe the traffic state of the road network, the improved KNN model uses the spatiotemporal matrix to describe the traffic state.

While each model has its advantages and can achieve good forecasting under different traffic conditions, It is difficult for a single model to cope with complex situations in the road network system. Zhou et al. proposed a deep learning-based multimodal learning framework to predict future traffic flow conditions based on current input data by dynamically selecting an optimal model or a subset of optimal models from a set of candidate models~\cite{zhou2019learning}. Based on the ensemble of Gradient Boosted Regression Tree (GBRT) and Least Absolute Shrinkage and Selection Operator (Lasso), Chen et al. proposed an ensemble learning algorithm for short-term traffic flow prediction, which takes into account accidents, accidents, adverse weather, work zones, and holidays~\cite{chen2019multi}. The previous multimodal traffic prediction models mostly considered the fusion of different models but seldom considered the impact of different data types on the results.

\subsection{Multimodal Representation Learning}

Since information of different modalities such as texts, figures, audios, videos are increasingly appearing in diverse applications, such as video classification~\cite{wu2016multistream}, object recognition tasks~\cite{song2016multimodal}, egocentric activity recognition~\cite{eitel2015multimodal}, etc., learning fusion representations of modalities has been going on for decades. There are three main types of methods to represent multimodal fusion. First, the joint representation, which integrates unimodal representations through mapping them together into a unified space. Based on joint representation, many classification or clustering tasks have been applied, such as event detection~\cite{wu2014zeroshot}~\cite{habibian2017video}, sentiment analysis~\cite{poria2016fusing}~\cite{zadeh2017tensor}, and visual question answering~\cite{fukui2016multimodal}. The second category is coordinated representation, which take advantages of cross-modal similarity models to represent unimodal in a coordinated space separately. The last one is encoder-decoder model, which aims to acquire intermediate layers that project one modality to another. This framework has been widely exploited for multimodal translation tasks, such as image caption~\cite{xu2015show}~\cite{vinyals2015show}, video description~\cite{donahue2017long}~\cite{venugopalan2014translating}, and image synthesis~\cite{reed2016generative}.

In recent years, deep learning based multimodal representation learning has received a lot of attention and research. Srivastava and Salakhutdinov~\cite{srivastava2012multimodal} proposed a Deep Boltzmann Machine model as a generative model, which can extract a unified representation that fuses modalities together by learning the joint space of image and text inputs. These raw inputs are transformed into corresponding high-level representations, then the model learns joint distribution of each representation. They found that if is useful for classification and information retrieval tasks from both unimodal and multimodal queries.

For multimodal data fusion, Ngiam et al.~\cite{ngiam2011multimodal} proposed Stacked AutoEncoder (SAE). The model not only gets a solution to cross-modality by obtaining better unimodal representation from knowledge of other modalities，but also learns the complicated correlation between modalities, with intermediate levels being learned as shared modality represent learning. To exploit the matching relationships between image and text modalities, Ma et al.~\cite{ma2015multimodal} proposed a multimodal convolutional neural network (m-CNN). By using an image CNN to grasp image content with a corresponding matching CNN, the joint representation of images and sentences is captured. To be specific, the matching CNN combines several semantic levels of words, from word level, to phase level, and sentence level. For image captions generation, Mao et al.~\cite{mao2014deep} proposed a multimodal recurrent neural architecture (m-RNN). The model contains two parts, one is deep recurrent neural networks for learning sentences and another is deep convolutional neural networks for understanding images. These two networks are in constant interaction with each other throughout the multimodal network.

In this study, spatial distributions from three different sources, including text and other inputs are learned and convolution operation is implemented on three modalities. The joint representation of sentences and numerical values are learned from 2D map space. Finally, these joint features are used to predict the classification of traffic congestion based on location query. Thus, the distribution of location query and multimodal representations are converged into a unified mapping space.

\begin{figure*}[t]
\centerline{\includegraphics[width= 1.0\textwidth]{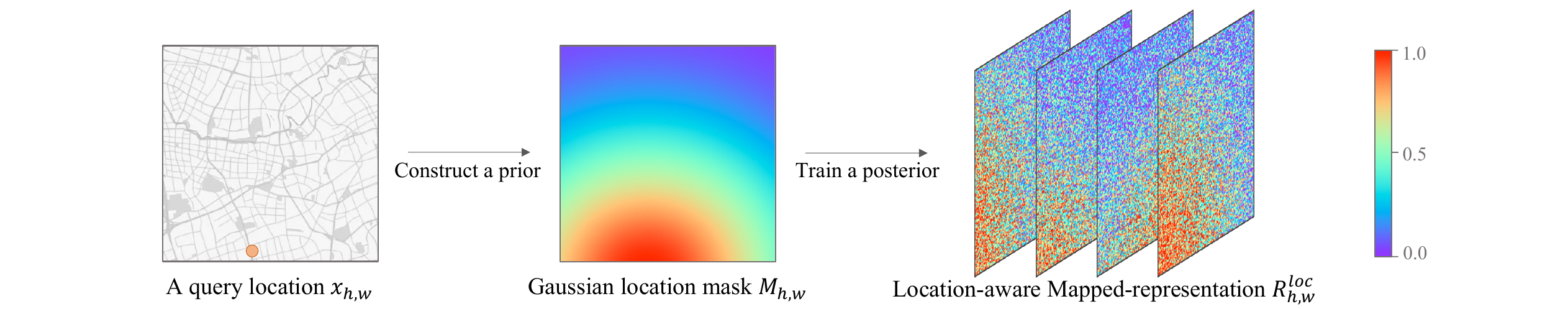}}
\caption{For a query location, a Gaussian location mask is constructed as a prior of the relations between the location and the whole map. Then, a Location-aware Mapped-representation (shown as a normalised heatmap) is optimised as a posterior by training the neural network in QLMM.}
\label{map}
\end{figure*}

\section{Methodology}

\subsection{Overview}

Current traffic prediction models are mostly based on deep learning. However, the processing of FNN is too time-consuming, while it is too difficult for RNNs to deal with high-resolution data. To step aside these limitations, a novel and lightweight end-to-end traffic situations prediction framework based on representation fusion is proposed, using CNNs to predict traffic congestion sequence at any location on a large-scale map efficiently, which employs multimodal social data with the location of the large-scale map as global reference information.

Specifically, the global reference information is divided into three categories (social media texts, real estates, and points of interest), which are scattered across the whole map with different distributions respectively (Fig.~\ref{data}). Also, being formatted learnedly and adaptively, the multimodal global reference information is learned and fused into a global-aware geo-preserving representation, also called meta-representation, for aggregating all reference information of the entire large-scale map. In addition, a geo-preserving representation is learned to match a specific location used to capture traffic congestion situations. Finally, the global-aware representation and the location-aware representation are aggregated and learned to obtain location-specific traffic congestion results.

\subsection{Problem Statement}

A traffic congestion prediction task can be formulated to utilise the proposed method to obtain valuable results. Firstly, the entire large-scale map is gridded to characterise all the data. As a result, the original map is formatted as a 2-dimensional grid array whose size is $H \times W$. For each grid on the map, three kinds of global reference information in the region are quantised as multi-dimensional features that are located in the correlated location of the whole gridded map, which is also regarded as three multi-channel matrices. 

Specifically, for the region represented by each grid, a vector with a length of $D_{mt}$ is obtained to represent the social media information from the dataset after pre-processing. For grids without any social media information, zero-vectors are used. Similarly, for a grid, pre-processed real estates information and points of interest information can be represented by a $D_{re}$-length vector and a $D_{pi}$-length vector respectively. As a result, for the entire gridded map, social media texts, real estates, and points of interest information can be formulated as three multi-channel matrices: $F_{mt}$, $F_{re}$, and $F_{pi}$, where $F_{mt} \in \mathbb{R}^{H \times W \times D_{mt}}$, $F_{re} \in \mathbb{R}^{H \times W \times D_{re}}$, and $F_{pi} \in \mathbb{R}^{H \times W \times D_{pi}}$, which are employed as a part of input of the proposed framework.

For the gridded map of size $H \times W$, different regions are divided into training, validation, and test regions, respectively, while the data in these regions are aggregated into training, validation, and test sets. With the help of reference information as well as traffic data in the training area, the task is to predict whether roads are congested in any unvisited grid area (test area). For the whole gridded map, the index of each grid $x_{h, w}$ is the location of the $h$th row and the $w$th column, where $h = 1, 2, ..., H$, and $w = 1, 2, ..., W$. For each $x_{h, w}$, the proposed model is able to obtain the results, $\hat y_{h, w}$, which is a vector with multi-length for describing whether the corresponding location is congested at multiple time points in a day. To this end, the proposed method is expected to minimise the difference between the predicted $\hat y_{h, w}$ and the ground truth of the location, which is denoted as $y_{h, w}$.

\subsection{Framework}

For the traffic congestion prediction task, the proposed novel and lightweight framework learns and generalises the global reference information well, so that the matching and fusion of specific query points are able to handle significant predictions. To be more specific, as shown in Fig.~\ref{framework}, featured global reference information ($F_{mt}$, $F_{re}$, and $F_{pi}$) of the whole large-scale map (Part. A in Fig.~\ref{framework}) is firstly fed into Multimodal Fusion and Generalisation Module (MFGM) (Part. B in Fig.~\ref{framework}) for obtaining the Global-aware Meta-representation (Part. C in Fig.~\ref{framework}) of the map. At the same time, a query location ($x_{h, w}$) is transferred into a Location-aware Mapped-representation (Part. E in Fig.~\ref{framework}) via Query Location Mapping Module (QLMM) (Part. D in Fig.~\ref{framework}). Finally, the Representation Fusion and Prediction Module (RFPM) (Part. E in Fig.~\ref{framework}) fuses the Global-aware Meta-representation and the Location-aware Mapped-representation and makes prediction via a neural network.

For each step in training, all features of the global reference information are always fed into the MFGM. As a result, when the entire model is trained to convergence, MFGM is robust enough to extract and generalise multimodal global reference information comprehensively and represent it as the Global-aware Meta-representation. Therefore, while inference, as an adapted representation of the whole map, a stable and informative Global-aware Meta-representation that is saved after training is directly fed into the RFPM without using the heavy MFGM architecture and the complex global reference information. In this way, the used framework maintains lightweight parameters and incredible efficiency.

\begin{figure*}[t]
\centerline{\includegraphics[width= 1.0\textwidth]{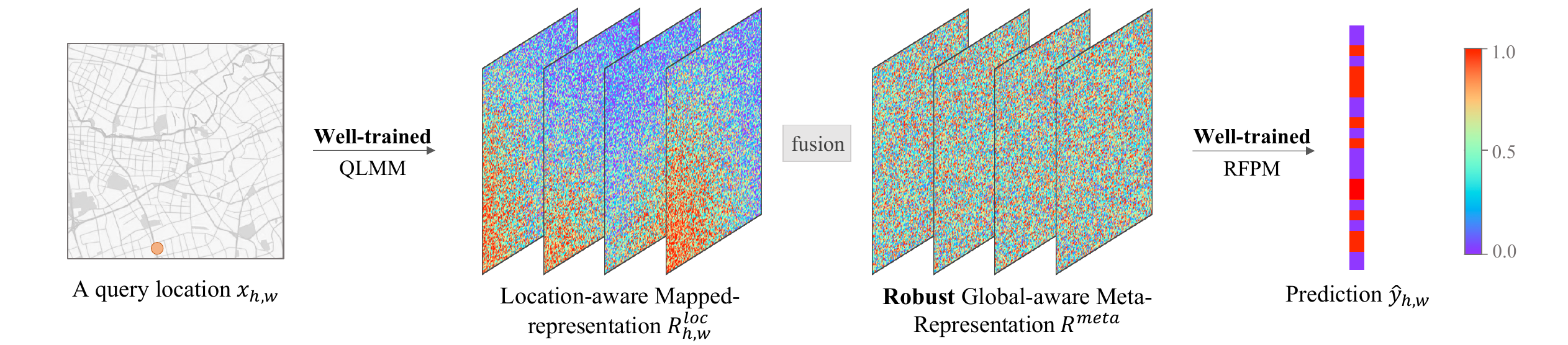}}
\caption{When training is completed, a Location-aware Mapped-representation (shown as a normalised heatmap) can be obtained via a well-trained QLMM. Then the saved robust Global-aware Meta-representation (shown as a normalised heatmap) is fused with generated Location-aware Mapped-representation for using as an input to make prediction via a well-trained RFPM.}
\label{inference}
\end{figure*}

The details of each module in the proposed framework will be introduced below:

\subsubsection{Multimodal Fusion and Generalisation Module (MFGM)}

For different types of global reference information which is formatted into the matrix with different channels, different convolutional neural networks are designed for obtaining three unimodal representations that have the same size, ${H_{ur} \times W_{ur} \times D_{ur}}$. As shown in Part. B of Fig.~\ref{framework}, there are three neural networks denoted as
$\mathcal{F}^{uni}=\left\{f_m^{uni}(\cdot)\right\}_{m=mt, re, pi}$ so that the obtained unimodal representations of three types of the global reference information is $\mathcal{U}=\left\{U_m\right\}_{m=mt, re, pi}$, where any $U_m \in \mathbb{R}^{H_{ur} \times W_{ur} \times D_{ur}}$, and the forward processing can be formulated as:

\begin{equation}
U_m = f_m^{uni}(F_m), \quad m = mt, re, pi.
\end{equation}
A channel-wise concatenation is achieved over the three unimodal representations causing a joint representation fed into a convolutional neural network, $f^{multi}(\cdot)$, similar to an Auto-Encoder. $f^{multi}(\cdot)$ mixes representations from different modalities and moulds them into a global reference-rich multi-channel representation with the number of channels $D_{meta}$ that preserves the spatial structure of the original map on each channel. The resulted representation is named as the Global-aware Meta-representation (Part. C in Fig.~\ref{framework}) and denoted as $R^{meta}$, where $R^{meta} \in \mathbb{R}^{H \times W \times D_{meta}}$, which is obtained by:

\begin{equation}
R^{meta} = f^{multi}(concat[U_{mt}, U_{re}, U_{pi}]).
\end{equation}

\subsubsection{Query Location Mapping Module (QLMM)}

For each query location in the entire gridded map, it is difficult to describe the spatial relationship between the location and the whole map well in the form of coordinate pairs ($x_{h,w}$). In contrast, generating a learnable geo-preserving representation is a reasonable approach for carrying the information of the location. More specifically, a location-rich multi-channel matrix that can adapt the location to the global information can better participate in the whole framework to help the neural network make predictions because the multi-channel matrix can spatially hold the interaction between the current query location and all other locations relative to the global information. The multi-channel matrix is named as Location-aware Mapped-representation (Part. E in Fig.~\ref{framework}) and denoted as $R_{h,w}^{loc}$ for each $x_{h,w}$.

However, with such complex high-dimensional interactions in geo-spatial, the distribution presented by the Location-aware Mapped-representation is not resolvable, therefore, a simple distribution is assumed as prior information of the Location-aware Mapped-representation. More precisely, in the initial state, the influence of the query location on the entire map spreads from the location to the surroundings in a two-dimensional Gaussian distribution. In other words, the probability that each location is affected by the query location follows a two-dimensional Gaussian distribution. Therefore, for each query location $x_{h,w}$, affected probability value of the grid at location $(h^{'},w^{'})$ is denoted as $m_{(h,w)\rightarrow(h^{'},w^{'})}^{'}$, and 

\begin{equation}
m_{(h,w)\rightarrow(h^{'},w^{'})}^{'} = P((a,b) = x_{h^{'},w^{'}}),
\end{equation}
where $(a,b)$ are two-dimensional random variables and $(a,b) \sim \mathcal{N}(x_{h,w}, \Sigma)$. As a result, for each query location $x_{h,w}$, a matrix of the same size as the original gridded map can be obtained as:

\begin{equation}
M_{h,w}^{'}=
\begin{bmatrix}
m_{(h,w)\rightarrow(1,1)}^{'} & \cdots   & m_{(h,w)\rightarrow(1,W)}^{'}  \\
\vdots & \ddots   & \vdots  \\
m_{(h,w)\rightarrow(H,1)}^{'} & \cdots\  & m_{(h,w)\rightarrow(H,W)}^{'}  \\
\end{bmatrix}.
\end{equation}
Then $M_{h,w}^{'}$ can be normalised for constructing a Gaussian location mask $M_{h,w}$ (shown in Fig.~\ref{map}). Based on the Gaussian location mask as a prior, the Location-aware Mapped-representation can be learned as a posterior via a convolutional neural network (shown in Fig.~\ref{map}), $f^{map}(\cdot)$ and the forward processing is shown in Part. D of Fig.~\ref{framework} and formulated as:

\begin{equation}
R_{h,w}^{loc} = f^{map}(M_{h,w}), \quad h \in [1, H], w \in [1, W].
\end{equation}

\subsubsection{Global-Location Representations}

As shown in Part. C and Part. E of Fig.~\ref{framework}, there are two key representations generated in the framework: Global-aware Meta-representation ($R^{meta}$) and Location-aware Mapped-representation ($R_{h,w}^{loc}$). In the proposed framework, Global-aware Meta-representation and Location-aware Mapped-representation are set to have the same size as the original gridded map for each channel. Also, these two representations have high-dimensional channels. In this way, for a query location, each grid has two high-dimensional vectors representing global reference data and spatial relations respectively, so that they can carry information for making predictions significantly.

Furthermore, the Global-aware Meta-representation is representing all of the global reference information, so that it can be utilised as a generalised feature map and has the potential to complete multiple tasks.

\subsubsection{Representation Fusion and Prediction Module (RFPM)}

A channel-wise fusion is implemented on Global-aware Meta-representation ($R^{meta}$) and Location-aware Mapped-representation ($R_{h,w}^{loc}$), which causes feeding for a neural network with a fully connected output layer, $f^{pred}(\cdot)$, in order to obtain the prediction results $\hat y_{h, w}$ for the query location $x_{h, w}$. The output $\hat y_{h, w}$ is a multi-dimensional vector whose single element represents the congestion at a time point of the monitored time series. The forward propagating is illustrated in Part. F of Fig.~\ref{framework} and formulated as:

\begin{equation}
\hat y_{h, w} = f^{pred}(concat[R^{meta}, R_{h,w}^{loc}]).
\end{equation}

\begin{figure*}[t]
\centerline{\includegraphics[width= 1.0\textwidth]{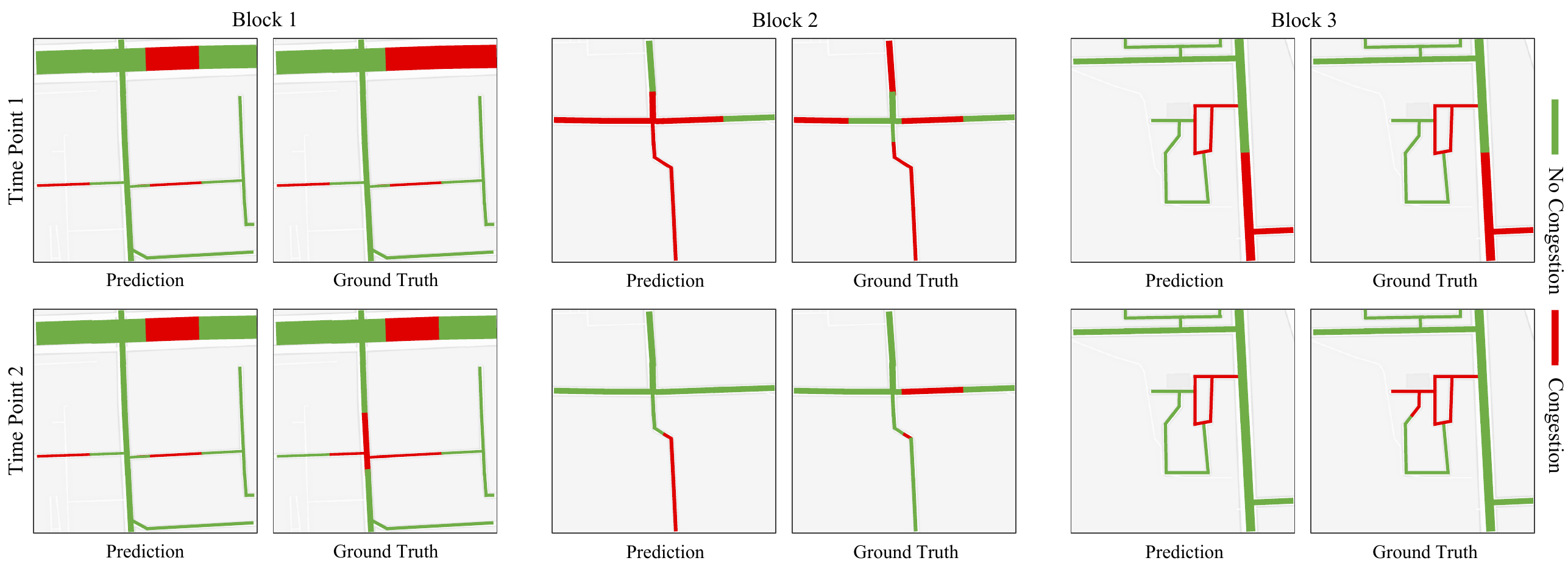}}
\caption{Qualitative results. The pairs of predicted and ground truth traffic congestion situations for three blocks at two time points are visualised respectively.}
\label{result}
\end{figure*}

\subsection{Optimisation and Efficient Inference}

When the model is trained, for each query location x, a temporal sequence of traffic congestion values $\hat y_{h, w}$ with $T$ time points is obtained, where $\hat y_{h, w} = [\hat y_{h, w}^{1}, \hat y_{h, w}^{2}, ..., \hat y_{h, w}^{T}]$, and the Mean Squared Error (MSE) is used as the objective function to measure the loss between $\hat y_{h, w}$ and the ground truth $y_{h, w}$, where $y_{h, w} = [y_{h, w}^{1}, y_{h, w}^{2}, ..., y_{h, w}^{T}]$. As a result, the loss can be formulated as:

\begin{equation}
Loss_{h, w} = \frac{1}{T} \sum_{t}^{T}\left\|\hat y_{h, w}^{t} - y_{h, w}^{t}\right\|_{2}^{2}.
\end{equation}
Based on the above objective function, the neural networks in MFGM, QLMM, and RFPM are trained to converge step by step. Since the global reference information is diverse and complex, MFGM is bulky and heavy. However, after the model converges, the Global-aware Meta-representation is preserved and can represent the massive global reference information stably. Therefore, when training is completed, only the lightweight neural networks in QLMM and RFPM are utilised so that the framework can achieve fast and efficient inference. More specifically, as shown in Fig.~\ref{inference}, the Location-aware Mapped-representation generated by QLMM is directly fused with the saved Global-aware Meta-representation and then fed into RFPM to complete the prediction.

\section{Experiments}

\subsection{Implementation details}

The experiments are mainly conducted on H4m dataset~\cite{zhao2022h4m, zhao2022pate}, which consists of $28, 550$ real estate, $497,256$ points of interest, $250,000$ traffic records and over $100$ million pieces of social media text geolocated in the same city. As for the different kinds of data, real estate, social media text, and points of interest are set as the global reference information. Also, the reference information is pre-processed for featuring it into multiple metrics. For each kind of reference information:

\begin{itemize}

\item \textbf{Social media texts:} All retrieved natural language texts sent out on social media are fed into a pre-trained BERT~\cite{devlin2018bert} to obtain a multi-dimensional vector of the grid with respect to social media texts. After the multi-dimensional vectors of all grids are obtained, a pre-trained Auto-Encoder~\cite{rumelhart1985learning} is used to reduce the dimensionality of the obtained vectors. Finally, the multi-channel matrix is obtained to be used as a part of the training data of the experiments.

\item \textbf{Real estates:} The average price of all real estates in each grid reflects the living quality of the area, so a normalised average real estates price is used as the real estates vector of the grid. And for grids without real estates information, zero is filled. Finally, for the entire gridded map, a single-channel matrix is obtained as real estates features.

\item \textbf{Points of interest:} There are 23 categories of points of interest in the grid, whose diversity constructs the rich interest information of the current grid. Therefore, count-based one-hot encoding is used to vectorise the interest points of the grid. As for the whole gridded map, points of interest features are represented as a multi-channel matrix for experiments.

\end{itemize}

At the same time, the traffic records are divided into training, validation, and testing set according to geographic locations, where the results on the testing set are reported. For more details, with in 250k grids on the whole map, there are around 24k grids total which are recorded with traffic data. For each grid, traffic data for 216 time points of one day is stored. All grids are divided into the training set, validation set, and test set at the ratio of 80\%, 20\%, and 20\%, respectively, with the fixed random seed in Sklearn. What is more, the traffic records are divided into two categories: congestion and no congestion, so that Accuracy, Recall, Precision, and F1-Score are used as metrics for evaluating the models. The code is available at: \url{https://github.com/luckkyzhou/TCP-MFRM}.

\begin{table}[t]
\caption{Main evaluation results and baseline results.}
\begin{center}
\begin{tabular}{c|cccc}
\hline
\multicolumn{1}{c|}{\textbf{Model}} &
\begin{tabular}[c]{@{}c@{}}\textbf{Accuracy}\\\textbf{($\%$)}\end{tabular} & \begin{tabular}[c]{@{}c@{}}\textbf{Recall}\\\textbf{($\%$)}\end{tabular} & \begin{tabular}[c]{@{}c@{}}\textbf{Precision}\\\textbf{($\%$)}\end{tabular} & \begin{tabular}[c]{@{}c@{}}\textbf{F1-Score}\\\textbf{($\%$)}\end{tabular}\\ \hline \hline
$Baseline_1$ & 56.16 & 48.16 & 53.21 & 50.56 \\
$Baseline_2$ & 57.84 & 49.85 & 55.21 & 52.40 \\
\textbf{$Ours$} & \textbf{65.60} & \textbf{58.98} & \textbf{64.15} & \textbf{61.46} \\
\hline
\end{tabular}
\label{main}
\end{center}
\end{table}

\subsection{Results}

\begin{table*}[t]
\caption{Evaluation results with different representations.}
\begin{center}
\begin{tabular}{c|cc|cccc}
\hline
\multicolumn{1}{c|}{\textbf{Architecture}} &
\begin{tabular}[c]{@{}c@{}}\textbf{Global-aware}\\\textbf{Meta-representation}\end{tabular} &
\begin{tabular}[c|]{@{}c@{}}\textbf{Location-aware }\\\textbf{Mapped-representation}\end{tabular} &
\begin{tabular}[c]{@{}l@{}}\textbf{Accuracy ($\%$)}\end{tabular} & \begin{tabular}[c]{@{}l@{}}\textbf{Recall ($\%$)}\end{tabular} & \begin{tabular}[c]{@{}l@{}}\textbf{Precision ($\%$)}\end{tabular} & \begin{tabular}[c]{@{}l@{}}\textbf{F1-Score ($\%$)}\end{tabular}\\ \hline \hline
\multirow{2}*{$Proposed$}
& & \checkmark & 62.53 & 52.73 & 61.28 & 56.68 \\
& \checkmark & & 53.48 & 46.46 & 49.97 & 48.15\\ \hline
\textbf{$Ours$} & \checkmark & \checkmark & \textbf{65.60} & \textbf{58.98} & \textbf{64.15} & \textbf{61.46}\\
\hline
\end{tabular}
\label{ablation1}
\end{center}
\end{table*}

\begin{table}[t]
\caption{Evaluation results on unimodal and multimodal inputs.}
\begin{center}
\begin{tabular}{c|cccc}
\hline
\multicolumn{1}{c|}{\textbf{Model}} &
\begin{tabular}[c]{@{}c@{}}\textbf{Accuracy}\\\textbf{($\%$)}\end{tabular} & \begin{tabular}[c]{@{}c@{}}\textbf{Recall}\\\textbf{($\%$)}\end{tabular} & \begin{tabular}[c]{@{}c@{}}\textbf{Precision}\\\textbf{($\%$)}\end{tabular} & \begin{tabular}[c]{@{}c@{}}\textbf{F1-Score}\\\textbf{($\%$)}\end{tabular}\\ \hline \hline
$Model_{mt}$ & 65.36 & 56.56 & \textbf{64.54} & 60.29 \\
$Model_{re}$ & 65.47 & 58.44 & 64.12 & 61.14 \\
$Model_{pi}$ & 65.32 & 58.32 & 63.93 & 61.00 \\ \hline
\textbf{$Ours$} & \textbf{65.60} & \textbf{58.98} & 64.15 & \textbf{61.46} \\
\hline
\end{tabular}
\label{ablation2}
\end{center}
\end{table}

After the entire framework is trained and validated on the selected dataset, quantitative results are obtained from the evaluation on the testing set. Also, a KNN-based traffic prediction method~\cite{lv2009real} is referenced as the baseline model. There are two different $k$ values used in baseline models ($k=3$ and $k=5$ are denoted as $Baseline_1$ and $Baseline_2$ respectively), and the quantitative results with the results of baseline are shown in Tab.~\ref{main}. At the same time, qualitative results for three blocks in the testing set at two time points are visualised in Fig.~\ref{result}. As can be seen from Tab.~\ref{main}, all metrics of the main experiment are better than the baseline, which adequately demonstrates the effectiveness of the proposed framework. As shown in Fig.~\ref{result}, the congestion of most roads is correctly predicted, regardless of whether it is a wide road or a narrow road.

According to the comparison between the baseline results and the main experimental results, it is difficult to obtain meaningful prediction results by simply learning and searching based on the simple geographical relationship between different roads on the large-scale map, which proves the necessity and effectiveness of the proposed framework.

\subsection{Ablation Study}

In order to demonstrate the effectiveness of the proposed modules, some experiments are designed by leaving parts of the whole framework unemployed while remaining the global structures. To investigate the role of the Global-aware Meta-representation and the Location-aware Mapped-representation on the effect of the overall framework, these two modules are used separately and evaluated to obtain the results. More specifically, for deactivated Global-aware Meta-representation, a multi-channel zero-matrix of equal size is used instead and fused with the Location-aware Mapped-representation; for the Location-aware Mapped-representation, a weak mapped-representation is obtained by using a one-hot location mask instead of the Gaussian location mask in QLMM. 

As can be seen from the evaluation results in Tab.~\ref{ablation1}, both Global-aware Meta-representation and Location-aware Mapped-representation are indispensable in the proposed framework. It is worth noting that the posterior information learned from the Gaussian location mask designed based on the Gaussian distribution plays a very important role, and at the same time, the reference information included in the Global-aware Meta-representation is also decisive in prediction.

At the same time, the effect of multimodal input on the global information aggregation of the framework is also tested ablatedly. For the unimodal representations generated by the three kinds of global reference information (social media texts, real estates, and points of interest), their independent active states (denoted as $Model_{mt}$, $Model_{re}$, and $Model_{pi}$) are used to evaluate the entire framework respectively. As shown in Tab.~\ref{ablation2}, all metrics under the fusion of multimodal representations are optimal except for secondly ranked precision, which indicates that multimodal global reference information with MFGM works better than any unimodal information for enabling the proposed framework to make significant predictions.

\section{Conclusion}

In this paper, an efficient convolutional neural network-based end-to-end framework is proposed, combining multimodal fusion and representation mapping to address the traffic congestion prediction problem on large-scale maps.

To achieve the expected prediction, a module called Multimodal Fusion and Generalisation Module (MFGM) is designed to aggregate global reference information to obtain the Global-aware Meta-representation for the entire map. At the same time, for a query location, a Location-aware Mapped-representation is obtained to describe the current query location relative to the global response via the Query Location Mapping Module (QLMM).

Finally, the Global-aware Meta-representation and the Location-aware Mapped-representation are fed into the Representation Fusion and Prediction Module (RFPM) and get traffic congestion predictions for the query location. Different contrast and ablation experiments are implemented on the proposed framework to verify its advancements and superiority.

{
\bibliographystyle{ieee_fullname}
\bibliography{main.bbl}

\begin{thebibliography}{10}\itemsep=-1pt

\bibitem{ahmed1979analysis}
Mohammed~S Ahmed and Allen~R Cook.
\newblock {\em Analysis of freeway traffic time-series data by using
  Box-Jenkins techniques}.
\newblock Number 722. 1979.

\bibitem{cai2016spatiotemporal}
Pinlong Cai, Yunpeng Wang, Guangquan Lu, Peng Chen, Chuan Ding, and Jianping
  Sun.
\newblock A spatiotemporal correlative k-nearest neighbor model for short-term
  traffic multistep forecasting.
\newblock {\em Transportation Research Part C: Emerging Technologies},
  62:21--34, 2016.

\bibitem{28}
Manoel Castro-Neto, Youngseon Jeong, Myong~K Jeong, and Lee~D Han.
\newblock Aadt prediction using support vector regression with data-dependent
  parameters.
\newblock {\em Expert Systems with Applications}, 36(2):2979--2986, 2009.

\bibitem{chan2011neural}
Kit~Yan Chan, Tharam~S Dillon, Jaipal Singh, and Elizabeth Chang.
\newblock Neural-network-based models for short-term traffic flow forecasting
  using a hybrid exponential smoothing and levenberg--marquardt algorithm.
\newblock {\em IEEE Transactions on Intelligent Transportation Systems},
  13(2):644--654, 2011.

\bibitem{chen2019multi}
Xiqun Chen, Shuaichao Zhang, and Li Li.
\newblock Multi-model ensemble for short-term traffic flow prediction under
  normal and abnormal conditions.
\newblock {\em IET Intelligent Transport Systems}, 13(2):260--268, 2019.

\bibitem{cheng2018deeptransport}
Xingyi Cheng, Ruiqing Zhang, Jie Zhou, and Wei Xu.
\newblock Deeptransport: Learning spatial-temporal dependency for traffic
  condition forecasting.
\newblock In {\em 2018 International Joint Conference on Neural Networks
  (IJCNN)}, pages 1--8. IEEE, 2018.

\bibitem{devlin2018bert}
Jacob Devlin, Ming-Wei Chang, Kenton Lee, and Kristina Toutanova.
\newblock Bert: Pre-training of deep bidirectional transformers for language
  understanding.
\newblock {\em arXiv preprint arXiv:1810.04805}, 2018.

\bibitem{donahue2017long}
Jeff Donahue, Lisa~Anne Hendricks, Marcus Rohrbach, Subhashini Venugopalan,
  Sergio Guadarrama, Kate Saenko, and Trevor Darrell.
\newblock Long-term recurrent convolutional networks for visual recognition and
  description.
\newblock {\em IEEE Transactions on Pattern Analysis and Machine Intelligence},
  39(4):677--691, 2017.

\bibitem{dougherty1993use}
Mark~S Dougherty, Howard~R Kirby, and Roger~D Boyle.
\newblock The use of neural networks to recognise and predict traffic
  congestion.
\newblock {\em Traffic engineering \& control}, 34(6), 1993.

\bibitem{eitel2015multimodal}
Andreas Eitel, Jost Springenberg, Luciano Spinello, Martin Riedmiller, and
  Wolfram Burgard.
\newblock Multimodal deep learning for robust rgb-d object recognition.
\newblock pages 681--687, 09 2015.

\bibitem{elfar2018machine}
Amr Elfar, Alireza Talebpour, and Hani~S Mahmassani.
\newblock Machine learning approach to short-term traffic congestion prediction
  in a connected environment.
\newblock {\em Transportation Research Record}, 2672(45):185--195, 2018.

\bibitem{fou2017}
Mohammadhani Fouladgar, Mostafa Parchami, Ramez Elmasri, and Amir Ghaderi.
\newblock Scalable deep traffic flow neural networks for urban traffic
  congestion prediction.
\newblock In {\em 2017 International Joint Conference on Neural Networks
  (IJCNN)}, pages 2251--2258. IEEE, 2017.

\bibitem{fukui2016multimodal}
Akira Fukui, Dong~Huk Park, Daylen Yang, Anna Rohrbach, Trevor Darrell, and
  Marcus Rohrbach.
\newblock Multimodal compact bilinear pooling for visual question answering and
  visual grounding.
\newblock In {\em EMNLP}, 2016.

\bibitem{habibian2017video}
Amirhossein Habibian, Thomas Mensink, and Cees G.~M. Snoek.
\newblock Video2vec embeddings recognize events when examples are scarce.
\newblock {\em IEEE Transactions on Pattern Analysis and Machine Intelligence},
  39(10):2089--2103, 2017.

\bibitem{karlaftis2011statistical}
Matthew~G Karlaftis and Eleni~I Vlahogianni.
\newblock Statistical methods versus neural networks in transportation
  research: Differences, similarities and some insights.
\newblock {\em Transportation Research Part C: Emerging Technologies},
  19(3):387--399, 2011.

\bibitem{levin1980forecasting}
Moshe Levin and Yen-Der Tsao.
\newblock On forecasting freeway occupancies and volumes (abridgment).
\newblock {\em Transportation Research Record}, (773), 1980.

\bibitem{li2018brief}
Yaguang Li and Cyrus Shahabi.
\newblock A brief overview of machine learning methods for short-term traffic
  forecasting and future directions.
\newblock {\em Sigspatial Special}, 10(1):3--9, 2018.

\bibitem{lv2009real}
Yisheng Lv, Shuming Tang, and Hongxia Zhao.
\newblock Real-time highway traffic accident prediction based on the k-nearest
  neighbor method.
\newblock In {\em 2009 international conference on measuring technology and
  mechatronics automation}, volume~3, pages 547--550. IEEE, 2009.

\bibitem{ma2015multimodal}
Lin Ma, Zhengdong Lu, Lifeng Shang, and Hang Li.
\newblock Multimodal convolutional neural networks for matching image and
  sentence.
\newblock 04 2015.

\bibitem{mao2014deep}
J. Mao, W. Xu, Y. Yang, J. Wang, Z. Huang, and A. Yuille.
\newblock Deep captioning with multimodal recurrent neural networks (m-rnn).
\newblock {\em arXiv preprint arXiv:1412.6632}, 2014.

\bibitem{ngiam2011multimodal}
Jiquan Ngiam, Aditya Khosla, Juhan Nam, Honglak Lee, and Andrew Ng.
\newblock Multimodal deep learning.
\newblock pages 689--696, 01 2011.

\bibitem{poria2016fusing}
Soujanya Poria, E. Cambria, Newton Howard, Guangbin Huang, and Amir Hussain.
\newblock Fusing audio, visual and textual clues for sentiment analysis from
  multimodal content.
\newblock {\em Neurocomputing}, 174:50--59, 2016.

\bibitem{reed2016generative}
Scott Reed, Zeynep Akata, Xinchen Yan, Lajanugen Logeswaran, Bernt Schiele, and
  Honglak Lee.
\newblock Generative adversarial text to image synthesis.
\newblock In Maria~Florina Balcan and Kilian~Q. Weinberger, editors, {\em
  Proceedings of The 33rd International Conference on Machine Learning},
  volume~48 of {\em Proceedings of Machine Learning Research}, pages
  1060--1069, New York, New York, USA, 20--22 Jun 2016. PMLR.

\bibitem{rumelhart1985learning}
David~E Rumelhart, Geoffrey~E Hinton, and Ronald~J Williams.
\newblock Learning internal representations by error propagation.
\newblock Technical report, California Univ San Diego La Jolla Inst for
  Cognitive Science, 1985.

\bibitem{34}
Bharti Sharma, Sachin Kumar, Prayag Tiwari, Pranay Yadav, and Marina~I
  Nezhurina.
\newblock Ann based short-term traffic flow forecasting in undivided two lane
  highway.
\newblock {\em Journal of Big Data}, 5(1):1--16, 2018.

\bibitem{song2016multimodal}
Sibo Song, Vijay Chandrasekhar, Bappaditya Mandal, Liyuan Li, Joo-Hwee Lim,
  Giduthuri~Sateesh Babu, Phyo~Phyo San, and Ngai-Man Cheung.
\newblock Multimodal multi-stream deep learning for egocentric activity
  recognition.
\newblock In {\em 2016 IEEE Conference on Computer Vision and Pattern
  Recognition Workshops (CVPRW)}, pages 378--385, 2016.

\bibitem{srivastava2012multimodal}
Nitish Srivastava and Russ~R Salakhutdinov.
\newblock Multimodal learning with deep boltzmann machines.
\newblock In F. Pereira, C.J. Burges, L. Bottou, and K.Q. Weinberger, editors,
  {\em Advances in Neural Information Processing Systems}, volume~25. Curran
  Associates, Inc., 2012.

\bibitem{32}
Shiliang Sun, Changshui Zhang, and Guoqiang Yu.
\newblock A bayesian network approach to traffic flow forecasting.
\newblock {\em IEEE Transactions on intelligent transportation systems},
  7(1):124--132, 2006.

\bibitem{ted2020}
David~Alexander Tedjopurnomo, Zhifeng Bao, Baihua Zheng, Farhana Choudhury, and
  AK Qin.
\newblock A survey on modern deep neural network for traffic prediction:
  Trends, methods and challenges.
\newblock {\em IEEE Transactions on Knowledge and Data Engineering}, 2020.

\bibitem{venugopalan2014translating}
Subhashini Venugopalan, Huijuan Xu, Jeff Donahue, Marcus Rohrbach, Raymond
  Mooney, and Kate Saenko.
\newblock Translating videos to natural language using deep recurrent neural
  networks.
\newblock 12 2014.

\bibitem{vinyals2015show}
Oriol Vinyals, Alexander Toshev, Samy Bengio, and Dumitru Erhan.
\newblock Show and tell: A neural image caption generator.
\newblock pages 3156--3164, 06 2015.

\bibitem{wu2014zeroshot}
Shuang Wu, Sravanthi Bondugula, Florian Luisier, Xiaodan Zhuang, and Pradeep
  Natarajan.
\newblock Zero-shot event detection using multi-modal fusion of weakly
  supervised concepts.
\newblock In {\em 2014 IEEE Conference on Computer Vision and Pattern
  Recognition}, pages 2665--2672, 2014.

\bibitem{wu2016multistream}
Zuxuan Wu, Yu-Gang Jiang, Xi Wang, Hao Ye, and Xiangyang Xue.
\newblock Multi-stream multi-class fusion of deep networks for video
  classification.
\newblock In {\em Proceedings of the 24th ACM International Conference on
  Multimedia}, MM '16, page 791–800, New York, NY, USA, 2016. Association for
  Computing Machinery.

\bibitem{xie2019sequential}
Zhipu Xie, Weifeng Lv, Shangfo Huang, Zhilong Lu, Bowen Du, and Runhe Huang.
\newblock Sequential graph neural network for urban road traffic speed
  prediction.
\newblock {\em IEEE Access}, 8:63349--63358, 2019.

\bibitem{xu2015show}
Kelvin Xu, Jimmy~Lei Ba, Ryan Kiros, Kyunghyun Cho, Aaron Courville, Ruslan
  Salakhutdinov, Richard~S. Zemel, and Yoshua Bengio.
\newblock Show, attend and tell: Neural image caption generation with visual
  attention.
\newblock In {\em Proceedings of the 32nd International Conference on
  International Conference on Machine Learning - Volume 37}, ICML'15, page
  2048–2057. JMLR.org, 2015.

\bibitem{27}
Lijin Yang, Qing Yang, Yonghua Li, and Yuqing Feng.
\newblock K-nearest neighbor model based short-term traffic flow prediction
  method.
\newblock In {\em 2019 18th International Symposium on Distributed Computing
  and Applications for Business Engineering and Science (DCABES)}, pages
  27--30. IEEE, 2019.

\bibitem{zadeh2017tensor}
Amir Zadeh, Minghai Chen, Soujanya Poria, Erik Cambria, and Louis-Philippe
  Morency.
\newblock Tensor fusion network for multimodal sentiment analysis.
\newblock In {\em Proceedings of the 2017 Conference on Empirical Methods in
  Natural Language Processing}, pages 1103--1114, Copenhagen, Denmark, sep
  2017. Association for Computational Linguistics.

\bibitem{zhao2022pate}
Yaping Zhao, Ramgopal Ravi, Shuhui Shi, Zhongrui Wang, Edmund~Y Lam, and
  Jichang Zhao.
\newblock Pate: Property, amenities, traffic and emotions coming together for
  real estate price prediction.
\newblock In {\em IEEE International Conference on Data Science and Advanced
  Analytics}. IEEE, 2022.

\bibitem{zhao2022h4m}
Yaping Zhao, Shuhui Shi, Ramgopal Ravi, Zhongrui Wang, Edmund~Y Lam, and
  Jichang Zhao.
\newblock H4m: Heterogeneous, multi-source, multi-modal, multi-view and
  multi-distributional dataset for socioeconomic analytics in case of beijing.
\newblock In {\em IEEE International Conference on Data Science and Advanced
  Analytics}. IEEE, 2022.

\bibitem{zheng2006}
Weizhong Zheng, Der-Horng Lee, and Qixin Shi.
\newblock Short-term freeway traffic flow prediction: Bayesian combined neural
  network approach.
\newblock {\em Journal of transportation engineering}, 132(2):114--121, 2006.

\bibitem{zhou2019learning}
Teng Zhou, Guoqiang Han, Xuemiao Xu, Chu Han, Yuchang Huang, and Jing Qin.
\newblock A learning-based multimodel integrated framework for dynamic traffic
  flow forecasting.
\newblock {\em Neural Processing Letters}, 49(1):407--430, 2019.

\end{thebibliography}
}

\end{document}